\documentclass[runningheads]{llncs}

\usepackage{graphicx}
\usepackage{comment}
\usepackage{diagbox}
\usepackage{ragged2e} 
\usepackage{array}
\newcolumntype{P}[1]{>{\centering\arraybackslash}m{#1}}
\newcolumntype{C}[1]{>{\centering\arraybackslash}m{#1}}
\usepackage[numbers]{natbib}
\usepackage{enumitem} 
\usepackage[colorlinks=true,linkcolor =blue, urlcolor=blue,citecolor = blue, breaklinks]{hyperref}

\newcommand{\textcite}[1]{\citet{#1}}
\newcommand{\parencite}[1]{\citep{#1}}

\begin{document}
\renewcommand\UrlFont{\color{blue}\rmfamily}
\title{Explainable Reinforcement Learning: A Survey}
%
%
\author{Erika Puiutta\inst{1}\orcidID{0000-0003-3796-8931}
\and Eric MSP Veith\inst{1}\orcidID{0000-0003-2487-7475}} 

\authorrunning{E. Puiutta \& E. Veith}
%
\institute{OFFIS -- Institute for Information Technology, Escherweg 2, 26121 Oldenburg, Germany
\email{erika.puiutta@offis.de, eric.veith@offis.de}\\}
%
\maketitle              
\begin{abstract}
Explainable Artificial Intelligence (XAI), i.e., the development of more transparent and interpretable AI models, has gained increased traction over the last few years. This is due to the fact that, in conjunction with their growth into powerful and ubiquitous tools, AI models exhibit one detrimential characteristic: a performance-trans\-pa\-rency trade-off. This describes the fact that the more complex a model's inner workings, the less clear it is how its predictions or decisions were achieved. But, especially considering Machine Learning (ML) methods like Reinforcement Learning (RL) where the system learns autonomously, the necessity to understand the underlying reasoning for their decisions becomes apparent. Since, to the best of our knowledge, there exists no single work offering an overview of Explainable Reinforcement Learning (XRL) methods, this survey attempts to address this gap. We give a short summary of the problem, a definition of important terms, and offer a classification and assessment of current XRL methods. We found that a) the majority of XRL methods function by mimicking and simplifying a complex model instead of designing an inherently simple one, and b) XRL (and XAI) methods often neglect to consider the human side of the equation, not taking into account research from related fields like psychology or philosophy. Thus, an interdisciplinary effort is needed to adapt the generated explanations to a (non-expert) human user in order to effectively progress in the field of XRL and XAI in general.

\end{abstract}
%
%
%
\section{Introduction}
\label{subsec:prob}
\begin{figure}[t]
\centering
\includegraphics[width=0.5\textwidth]{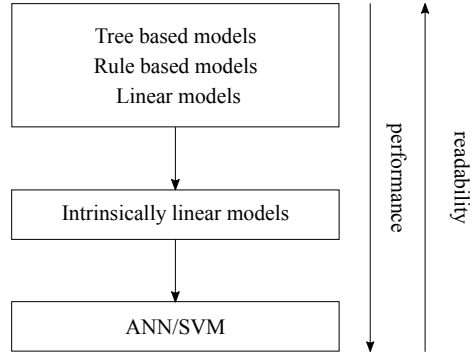}
\caption{Schematic representation of the performance-readability trade-off. Simpler, linear models are easy to understand and interpret, but suffer from a lack of performance, while non-linear, more flexible models are too complex to be understood easily. Adopted from \citeauthor{Martens2011} \cite{Martens2011}.} \label{tradeoff}
\end{figure}

Over the past decades, AI has become ubiquitous in many areas of our everyday lives. Especially Machine Learning (ML) as one branch of AI has numerous fields of application, be it transportation \cite{Tomzcak2019}, advertisement and content recommendation \cite{Nguyen2014}, or  medicine \cite{LiuCancer}. Unfortunately, the more powerful and flexible those models are, the more opaque they become, essentially making them black boxes (see figure \hyperref[tradeoff]{\ref{tradeoff}}). This trade-off is referred to by different terms in the literature, e.g. readability-performance trade-off \cite{Dosilovic2018}, accuracy-comprehensibility trade-off \cite{Freitas2014}, or accuracy-interpretability trade-off \cite{Rudin2019}. This work aims to, first, establish the need for explainable AI in general and explainable RL specifically. After that, the general concept of RL is briefly explained and the most important terms related to XAI are defined. Then, a classification of XAI models is presented and selected XRL models are sorted into these categories. Since there already is an abundance of sources on XAI but less so about XRL specifically, the focus of this work lies on providing information about and presenting sample methods of XRL models\footnote{Please note that, while there is a distinction between Reinforcement Learning and Deep Reinforcement Learning (DRL), for the sake of simplicity, we will refer to both as just Reinforcement Learning going forward.}. Thus, we present one method for each category in more detail and give a critical evaluation over the existing XRL methods.

\subsection{The importance of explainability}
Why is explainability so crucial? First, there is one obvious psychology-related reason: `if the users do not trust a model or a prediction, they will not use it' \cite[p. 1]{Ribeiro2016}. Trust is an essential prerequisite of using a model or system \cite{Israelsen2019,Dosilovic2018}, and transparency has been identified as one key component both in increasing users' trust \cite{Glass2008}, as well as users' acceptance of a system \cite{Herlocker2000} (for a formal definition of transparency and related terms, see section \hyperref[subsec:defs]{\ref{subsec:defs}}). Transparency also justifies a system's decisions and enables them to be fair and ethical \cite{Adadi2018}. Thus, in order to confidently use a system, it needs to be trusted, and in order to be trusted, it needs to be transparent and its decisions need to be justifiable.

Second, AI technologies have become an essential part in almost all domains of Cyber-Physical Systems (CPSs). Reasons include the thrive for increased efficiency, business model innovations, or the necessity to accommodate volatile parts of today's critical infrastructures, such as a high share of renewable energy sources. In time, AI technologies evolved from being an additional input to an otherwise soundly defined control system, to increasing the state awareness of a CPS---e.g., \emph{Neural State Estimation}~\parencite{dehghanpour2018survey}---to fully decentralized, but still rule-governed systems---such as the \emph{Universal Smart Grid Agent}~\parencite{veith2017universal}---, to a system where all behavior originates from machine learning. \emph{AlphaGo}, \emph{AlphaGo Zero}, and \emph{MuZero} are probably being the most widely-known representatives of the last category~\parencite{silver2017mastering,schrittwieser2019mastering}, but for CPS analysis and operation, Adversarial Resilience Learning (ARL) has emerged as a novel methodology based on DRL \parencite{Fischer2019arl,veith2019cpsanalysis}. It is specifically designed to analyse and control critical infrastructures; obviously, explainability is tantamount here.

There is also a legal component to be considered; the EU General Data Protection Regulation (GDPR) \cite{EUGDPR}, which came into effect in May 2018, aims to ensure a `right to explanation' \cite[p. 1]{Goodman2017} concerning automated decision-making and profiling. It states that `[...] such processing should subject to suitable safeguards, which should include [...] the right to obtain human intervention [...] [and] an explanation of the decision reached after such assessment' \cite[recital 71]{EUGDPR}. Additionally, the European Commission set out an AI strategy with transparency and accountability as important principles to be respected \cite{COM2018}, and in their Guidelines on trustworthy AI \cite{COM2019} they state seven key requirements, with transparency and accountability as two of them.

Finally, there are important practical reasons to consider; despite the increasing efficiency and versatility of AI, its incomprehensibility reduces its usefulness, since `incomprehensible decision-making can still be effective, but its effectiveness does not mean that it cannot be faulty' \cite[p. 1]{Lee2019}. For example, in \cite{Szegedy2013}, neural nets successfully learnt to classify pictures but could be led to misclassification by (to humans) nearly imperceptible perturbations, and in \cite{Nguyen_2015_CVPR}, deep neural nets classified unrecognizable images with \textgreater99\% certainty. This shows that a high level of effectiveness (under standard conditions) or even confidence does not imply that the decisions are correct or based on appropriately-learnt data.

Bearing this in mind, and considering the fact that, nowadays, AI can act increasingly autonomous, explaining and justifying the decisions is now more crucial than ever, especially in the domain of RL where an agent learns by itself, without human interaction.

\subsection{Reinforcement Learning}
Reinforcement Learning is a trial-and-error learning algorithm in which an autonomous agent tries to find the optimal solution to a problem through automated learning \cite{sequeirainterestinglong}. Possible applications for the use of RL are teaching neural networks to play games like Go \cite{silver2017mastering}, teaching robots to perform certain tasks \cite{Kober2013}, or intelligent transport systems \cite{Arel2010}.
RL is usually introduced as a Markov Decision Process (MDP) if it satisfies the Markov property: the next state depends only on the current state and the agent's action(s), not on past states \cite{kaelbling1996reinforcement}\footnote{To be exact, MDPs assume that the complete world state is visible to the agent which is, naturally, not always true. In these cases, a \emph{partially observable Markov decision process} (POMDP) can be used where, instead of observing the current state directly, we have a probability distribution over the possible states instead \cite{POMDP}.
For the sake of simplicity,  we do not go into further detail and refer the reader to \citeauthor{Kaelbling1998} or \citeauthor{kimura1997reinforcement} \cite{Kaelbling1998,kimura1997reinforcement} for more information.}.

\begin{figure}[t]
    \centering
    \includegraphics[width=0.6\textwidth]{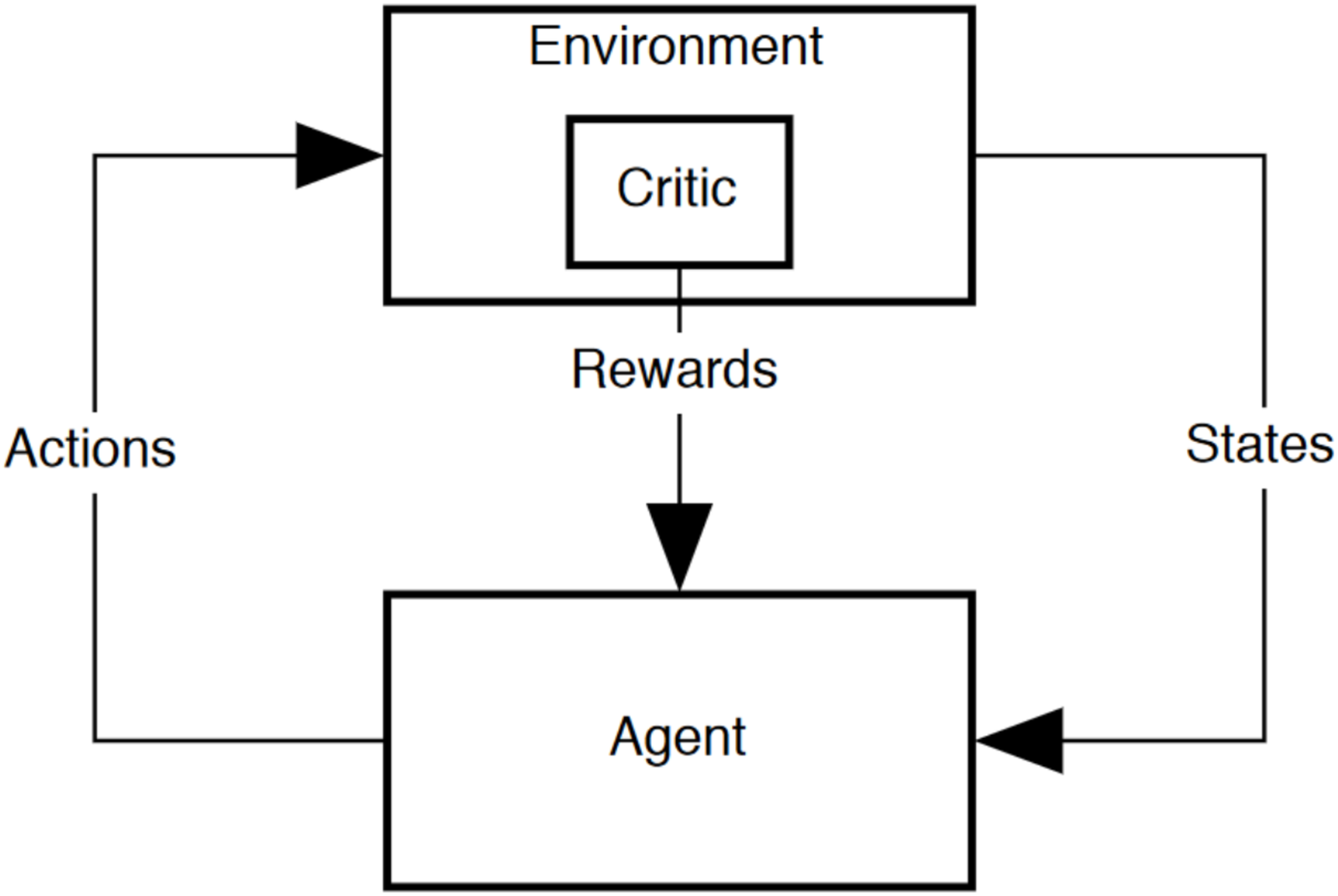}
    \caption{Interaction between agent and environment in RL. The agent performs a certain action which is rewarded by the `critic' in the environment, and it receives an update on the environment's states. Adapted from \citeauthor{barto2004intrinsically} \cite{barto2004intrinsically}}
    \label{fig:RLSchema}
\end{figure}

The learning process is initiated by an agent randomly performing an action which leads to a certain environmental state. This state has a reward assigned to it depending on how desirable this outcome is, set by the designer of the task (see also figure \hyperref[fig:RLSchema]{\ref{fig:RLSchema}}). The algorithm will then learn a policy, i.e., an action-state-relation, in order to maximize the cumulative reward and be able to select the most optimal action in each situation. For more information on RL, see also \cite{sequeirainterestinglong,LiDeepRL}.

\subsection{Definition of important terms}
\label{subsec:defs}
As already mentioned in section \hyperref[subsec:prob]{\ref{subsec:prob}}, the more complex a systems becomes, the less obvious its inner workings become. Additionally, there is no uniform term for this trade-off in the literature; XAI methods use an abundance of related, but distinct terms like transparency, reachability, etc... This inconsistency can be due to one or both of the following reasons: a) different terms are used in the same sense due to a lack of official definition of these terms, or b) different terms are used because the authors (subjectively) draw a distinction between them, without an official accounting of these differences. In any case, a uniform understanding and definition of what it means if a method is described as `interpretable' or `transparent' is important in order to clarify the potential, capacity and intention of a model. This is not an easy task, since there is no unique definition for the different terms to be found in the literature; even for `interpretability', the concept which is most commonly used, `the term [...] holds no agreed upon meaning, and yet machine learning conferences frequently publish papers which wield the term in a quasi-mathematical way' \cite{lipton2016zitat}. In \citeauthor{doshivelez2017rigorous} \cite[p. 2]{doshivelez2017rigorous}, interpretability is 'the ability to explain or to present in understandable terms to a human', however, according to \citeauthor{Kim2016examples} \cite[p. 7]{Kim2016examples} `a method is interpretable if a user can correctly and efficiently predict the method’s result'. Some authors use transparency as a synonym for interpretability \cite{lipton2016zitat}, some use comprehensibility as a synonym \cite{Freitas2014}, then again others draw a distinction between the two \cite{doran2017does} (for more information on how the different terms are used in the literature, we refer the reader to \cite{lipton2016zitat,Lipton2018,doshivelez2017rigorous,Freitas2014,Kim2016examples,doran2017does,Montavon2018,Chakraborty2017}). If we tackle this issue in a more fundamental way, we can look at the definition of `to interpret' or `interpretation'. The Oxford Learners Dictionary\footnote{https://www.oxfordlearnersdictionaries.com/} defines it as follows:
\begin{itemize}
             \item[$\bullet$] to explain the meaning of something
             \item[$\bullet$] to decide that something has a particular meaning and to understand it in this way
             \item[$\bullet$] to translate one language into another as it is spoken
             \item[$\bullet$] the particular way in which something is understood or explained
\end{itemize}
Seeing that, according to the definition, interpretation contains an explanation, we can look at the definition for `to explain'/`explanation':
\begin{itemize}
             \item[$\bullet$] to tell somebody about something in a way that makes it easy to understand
             \item[$\bullet$] to give a reason, or be a reason, for something
             \item[$\bullet$] a statement, fact, or situation that tells you why something happened
             \item[$\bullet$] a statement or piece of writing that tells you how something works or makes something easier to understand
\end{itemize}
Both definitions share the notion of conveying the reason and meaning of something in order to make someone understand, but while an explanation is focused on \emph{what} to explain, an interpretation has the additional value of considering \emph{how} to explain something; it translates and conveys the information in a way that is more easily understood.
And that is, in our opinion, essential in the frame of XAI/XRL: not only extracting the necessary information, but also presenting it in an appropriate manner, translating it from the `raw data' into something humans and especially laypersons can understand.


So, because we deem a shared consensus on the nomenclature important, we suggest the use of this one uniform term, \emph{interpretability}, to refer to the ability to not only extract or generate explanations for the decisions of the model, but also to present this information in a way that is understandable by human (non-expert) users to, ultimately, enable them to predict a model's behaviour.



%

\section{XAI Taxonomy}
\label{sec:tax}
\begin{figure}[t]
\centering
\includegraphics[width=0.7\textwidth]{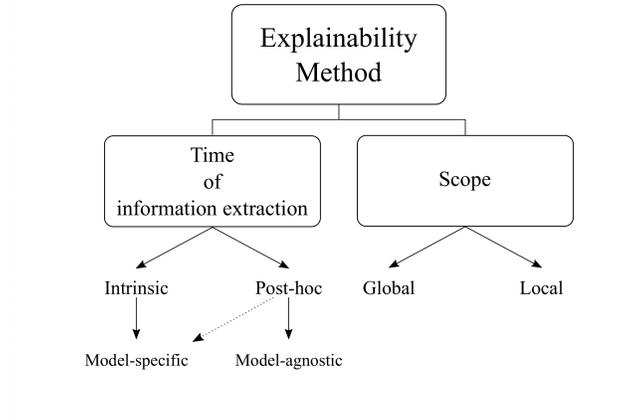}
\caption{A pseudo ontology of XAI methods taxonomy. Adapted from \citeauthor{Adadi2018} \cite{Adadi2018}.}
\label{globallocal}
\end{figure}

\begin{figure}[t]
\centering
\includegraphics[width=0.85\textwidth]{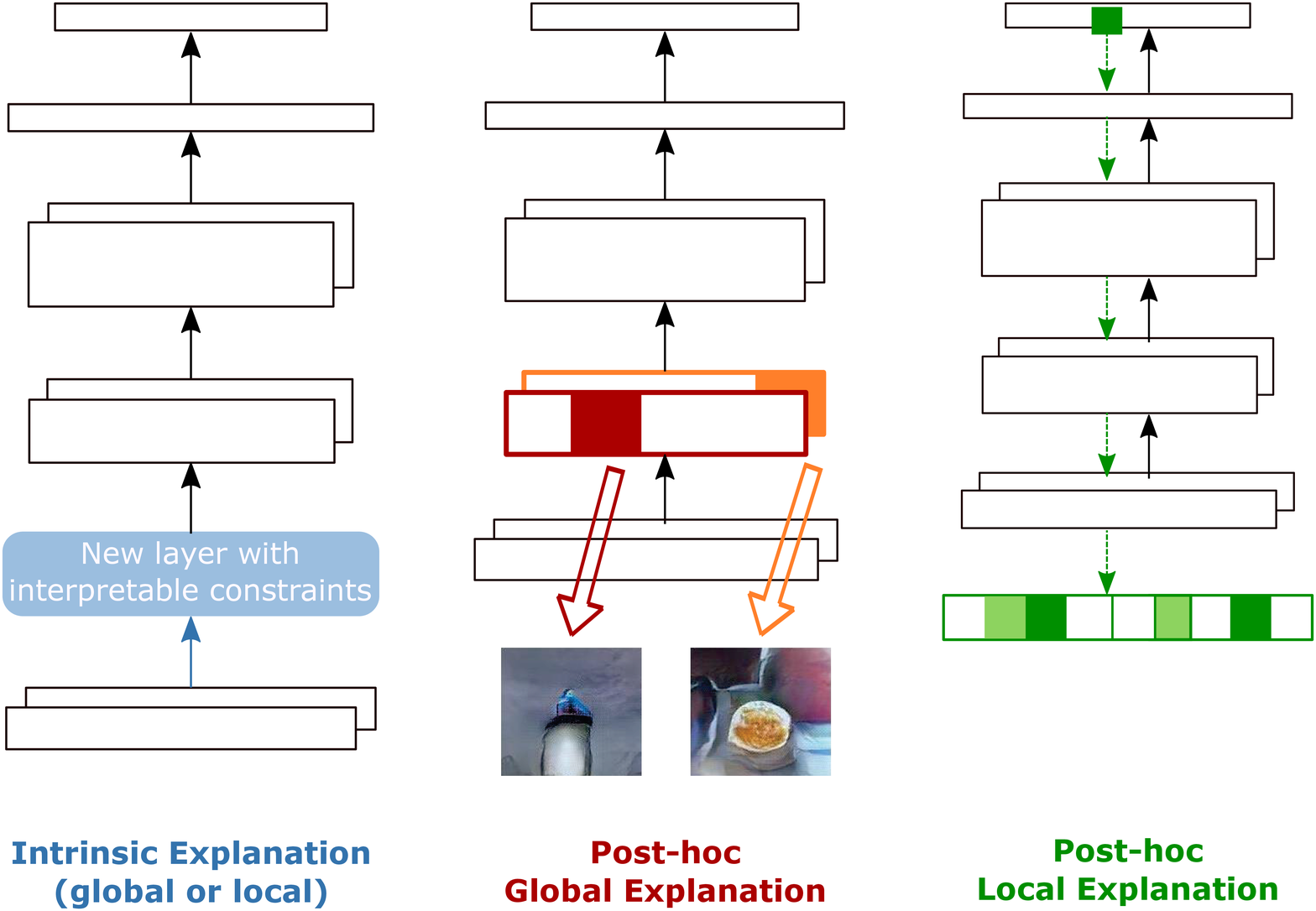}
\caption{An illustration of global vs. local, and intrinsic vs. post-hoc interpretable machine learning techniques, with a deep neural network as an example. On the left, the model and the layers' constraints are built in a way that is inherently interpretable (intrinsic interpretability). The middle and right column show post-hoc interpretability, achieved by a global and local explanation, respectively. The global explanation explains the different representations corresponding to the different layers in general, while the local explanation illustrates the contribution of the different input features to a certain output. Adopted from \citeauthor{Du2019} \cite{Du2019}.}
\label{globallocalacm}
\end{figure}

XAI methods can be categorized based on two factors; first, based on when the information is extracted, the method can be intrinsic or post-hoc, and second, the scope can be either global or local (see figure \hyperref[globallocal]{\ref{globallocal}}, and figure \hyperref[globallocalacm]{\ref{globallocalacm}} for examples). 

Global and local interpretability refer to the scope of the explanation; global models explain the entire, general model behaviour, while local models offer explanations for a specific decision \cite{githubtypeintr}. Global models try to explain the whole logic of a model by inspecting the structures of the model \cite{Adadi2018,Du2019}. Local explanations try to answer the question: `Why did the model make a certain prediction/decision for an instance/for a group of instances?' \cite{githubtypeintr,Adadi2018}. They also try to identify the contributions of each feature in the input towards a specific output \cite{Du2019}. Additionally, global interpretability techniques lead to users trusting a model, while local techniques lead to trusting a prediction \cite{Du2019}.

Intrinsic vs. post-hoc interpretability depend on the time when the explanation is extracted/generated; An intrinsic model is a ML model that is constructed to be inherently interpretable or self-explanatory at the time of training by restricting the complexity of the model \cite{Du2019}. Decision trees, for example, have a simple structure and can be easily understood \cite{githubtypeintr}. Post-hoc interpretability, in contrast, is achieved by analyzing the model after training by creating a second, simpler model, to provide explanations for the original model \cite{Du2019,githubtypeintr}. Surrogate models or saliency maps are examples for this type \cite{Adadi2018}. Post-hoc interpretation models can be applied to intrinsic interpretation models, but not necessarily vice versa. Just like the models themselves, these interpretability models also suffer from a transparency-accuracy-trade-off; intrinsic models usually offer accurate explanations, but, due to their simplicity, their prediction performance suffers. Post-hoc interpretability models, in contrast, usually keep the accuracy of the original model intact, but are harder to derive satisfying and simple explanations from \cite{Du2019}.

Another distinction, which usually coincides with the classification into intrinsic and post-hoc interpretability, is the classification into model-specific or model-agnostic. Techniques are model-specific if they are limited to a specific model or model class \cite{githubtypeintr}, and they are model-agnostic if they can be used on any model \cite{githubtypeintr}. As you can also see in figure \hyperref[globallocal]{\ref{globallocal}}, intrinsic models are model-specific, while post-hoc interpretability models are usually model-agnostic.

\citeauthor{Adadi2018} \cite{Adadi2018} offer an overview of common explainability techniques and their rough (i.e., neither mutually exclusive nor exhaustive) classifications into these categories. In section \hyperref[sec:listmeth]{\ref{sec:listmeth}}, we follow their example and provide classifications for a list of selected XRL method papers.


\section{Non-exhaustive list of XRL methods}
\label{sec:listmeth}
A literature review was conducted using the database Google Scholar. Certain combinations of keywords were used to select papers; first, `explainable reinforcement learning', and `XRL' together with `reinforcement learning' and `machine learning' were used. Then, we substituted `explainable' for common variations used in literature like `explainable', `transparent', and `understandable'. We then scanned the papers for relevance and consulted their citations and reference lists for additional papers. Because we only wanted to focus on current methods, we restricted the search to papers from 2010-2020. Table \hyperref[tab:einordnung]{\ref{tab:einordnung}} shows the list of selected papers and their classification according to section \hyperref[sec:tax]{\ref{sec:tax}} based on our understanding.

For a more extensive demonstration of the different approaches, we chose the latest paper of each quadrant \footnote{With the exception of method C in section \hyperref[subsec:methc]{\ref{subsec:methc}} where we present a Linear Model U-Tree method although another paper with a different, but related method was published slightly later. See the last paragraph of that section for our reasoning for this decision.} and explain them in more detail in the following sections as an example for the different XRL methods. 

\bgroup
\def\arraystretch{1}%
\setlength\tabcolsep{0.2cm}
\begin{table}[t]
    \centering
    \caption{Selected XRL methods and their categorization according to the taxonomy described in section \hyperref[sec:tax]{\ref{sec:tax}}.}
    \begin{tabular}{C{1.5cm}|C{4.9cm}|C{4.9cm}}
         \diagbox[width=1.9cm]{Time}{Scope} & Global & Local  \\
         \hline
        Intrinsic & \begin{itemize}[leftmargin=*, rightmargin=0.2cm]
             \item[$\bullet$] \textbf{PIRL} (\citeauthor{PIRL} \cite{PIRL})
              \item[$\bullet$] Fuzzy RL policies (\citeauthor{Hein2017} \cite{Hein2017})
         \end{itemize} & \begin{itemize}[leftmargin=*, rightmargin=0.2cm]
             \item[$\bullet$] \textbf{Hierarchical Policies} (\citeauthor{minecraft} \cite{minecraft})
         \end{itemize} \\ \hline
         Post-hoc & \begin{itemize}[leftmargin=*, rightmargin=0.2cm]
             \item[$\bullet$] Genetic Programming (\citeauthor{Hein2018} \cite{Hein2018})
             \item[$\bullet$]  Reward Decomposition (\citeauthor{juozapaitis2019explainable} \cite{juozapaitis2019explainable})
             \item[$\bullet$] Expected Consequences (\citeauthor{vanderwaacontr} \cite{vanderwaacontr})
             \item[$\bullet$] Soft Decision Trees (\citeauthor{coppens2019distilling} \cite{coppens2019distilling})
             \item[$\bullet$] Deep Q-Networks (\citeauthor{zahavy2016} \cite{zahavy2016})
             \item[$\bullet$] Autonomous Policy Explanation (\citeauthor{Hayes2017} \cite{Hayes2017})
             \item[$\bullet$] Policy Distillation (\citeauthor{andrei2015} \cite{andrei2015})
             \item[$\bullet$] \textbf{Linear Model U-Trees} (\citeauthor{Liu2019utree} \cite{Liu2019utree})
             \end{itemize} &  \begin{itemize}[leftmargin=*, rightmargin=0.2cm]
             \item[$\bullet$] Interestingness Elements (\citeauthor{sequeirainterestinglong} \cite{sequeirainterestinglong})
             \item[$\bullet$] Autonomous Self-Explanation (\citeauthor{Fukuchi2017} \cite{Fukuchi2017})
             \item[$\bullet$] \textbf{Structural Causal Model} (\citeauthor{MadumalCausal} \cite{MadumalCausal})
             \item[$\bullet$] Complementary RL (\citeauthor{Lee2019} \cite{Lee2019})
             \item[$\bullet$] Expected Consequences (\citeauthor{vanderwaacontr} \cite{vanderwaacontr})
             \item[$\bullet$] Soft Decision Trees (\citeauthor{coppens2019distilling} \cite{coppens2019distilling})
             \item[$\bullet$] \textbf{Linear Model U-Trees} (\citeauthor{Liu2019utree} \cite{Liu2019utree})
             \end{itemize} \\ 
    \end{tabular}
    {\raggedright \textbf{Notes.} Methods in bold are presented in detail in this work. \par}
    \label{tab:einordnung}
\end{table}

\subsection{Method A: Programmatically Interpretable Reinforcement Learning}
\label{subsec:metha}
\citeauthor{PIRL} \cite{PIRL} have developed `PIRL', a Programmatically Interpretable Reinforcement Learning framework, as an alternative to DRL. In DRL, the policies are represented by neural networks, making them very hard (if not impossible) to interpret. The policies in PIRL, on the other hand, while still mimicking the ones from the DRL model, are represented using a high-level, human-readable programming language. Here, the problem stays the same as in traditional RL (i.e., finding a policy that maximises the long-term reward), but in addition, they restrict the vast amount of target policies with the help of a \emph{(policy) sketch}. To find these policies, they employ a framework which was inspired by imitation learning, called \emph{Neurally Directed Program Search (NDPS)}. This framework first uses DRL to compute a policy which is used as a neural `oracle' to direct the policy search for a policy that is as close as possible to the neural oracle. Doing this, the performances of the resulting policies are not as high than the ones from the DRL, but they are still satisfactory and, additionally, more easily interpretable.
They evaluate this framework by comparing its performance with, among others, a traditional DRL framework in The Open Racing Car Simulator (TORCS) \cite{wymann2000torcs}. Here, the controller has to set five parameters (acceleration, brake, clutch, gear and steering of the car) to steer a car around a race track as fast as possible. Their results show that, while the DRL leads to quicker lap time, the NDPS still outperforms this for several reasons: it shows much smoother driving (i.e., less steering actions) and is less perturbed by noise and blocked sensors. It also is easier to interpret and is better at generalization, i.e., it performs better in situations (in this case, tracks) not encountered during training than a DRL model.
Concerning restrictions of this method, it is worth noting that the authors only considered environments with symbolic inputs, not perceptual, in their experiments. They also only considered deterministic policies, not stochastic policies.

\subsection{Method B: Hierarchical and Interpretable Skill Acquisition in Multi-task Reinforcement Learning}
\label{subsec:methb}
\citeauthor{minecraft}\cite{minecraft} proposed a new framework for multi-task RL using hierarchical policies that addressed the issue of solving complex tasks that require different skills and are composed of several (simpler) subtasks. It is based on and extends multi-task RL with modular policy design through a two-layer hierarchical policy \cite{andreas2017} by incorporating less assumptions, and, thus, less restrictions. They trained and evaluated their model with object manipulation tasks in a Minecraft game setting (e.g. finding, getting, or stacking blocks of a certain color), employing advantage actor-critic as policy optimization using off-policy learning. The model is hierarchical because each top-level policy (e.g., `stack x') consists of several lower levels of actions (`find x'$\rightarrow$ `get x'$\rightarrow$ `put x', see also figure \hyperref[fig:hierarc]{\ref{fig:hierarc}}). The novelty of this method is the fact that each task is described by a human instruction (e.g. `stack blue'), and agents can only access learnt skills through these descriptions, making its policies and decisions inherently human-interpretable.

Additionally, a key idea of their framework is that a complex task could be decomposed into several simpler subtasks. If these sub-tasks could be fulfilled by employing an already learnt `base policy', no new skill had to be learnt; otherwise, it would learn a new skill and perform a different, novel action. To boost efficiency and accuracy, the framework also incorporated a stochastic temporal grammar model that was used to model temporal relationships and priorities of tasks (e.g., before stacking a block on top of another block, you must first obtain said block).

The resulting framework could efficiently learn hierarchical policies and representations in multi-task RL, only needing weak human supervision during training to decide which skills to learn. Compared to a flat policy that directly maps the state and instruction to an action, the hierarchical model showed a higher learning efficiency, could generalize well in new environments, and was inherently interpretable. 

\begin{figure}[t]
    \centering
    \includegraphics[width=\textwidth]{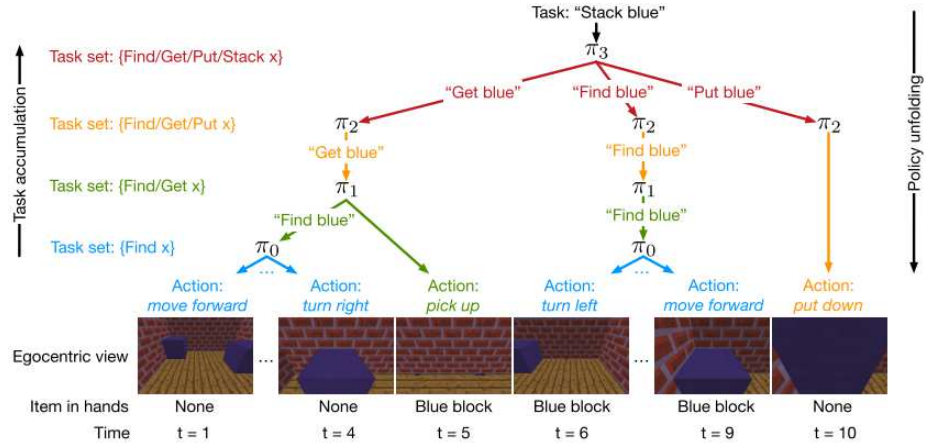}
    \caption{Example for the mutli-level hierarchical policy for the task to stack two blue boxes on top of each other. The top-level policy ($\pi_{3}$, in red) encompasses the high-level plan `get blue'$\rightarrow$`find blue'$\rightarrow$`put blue'. Each step (i.e., arrow) either initiates another policy (marked by a different color) or directly executes an action. Adopted from \cite{minecraft}.}
    \label{fig:hierarc}
\end{figure}

\subsection{Method C: Toward Interpretable Deep Reinforcement Learning with Linear Model U-Trees}
\label{subsec:methc}
In \citeauthor{Liu2019utree} \cite{Liu2019utree}, a mimic learning framework based on stochastic gradient descent is introduced. This framework approximates the predictions of an accurate, but complex model by mimicking the model's Q-function using Linear Model U-Trees (LMUTs). LMUTs are an extension of Continuous U-Trees (CUTs) which were developed to approximate continuous functions \cite{uther1998tree}. The difference between CUTs and LMUTs is that, instead of constants, LMUTs have a linear model at each leaf node which also improves its generalization ability. They also generally have fewer leaves and are therefore simpler and more easily understandable. The novelty of this method lies in the fact that other tree representations used for interpretations were only developed for supervised learning, not for DRL.

The framework can be used to analyze the importance of input features, extract rules, and calculate `super-pixels' (`contiguous patch[es] of similar pixels' \cite[p. 1]{Ribeiro2016}) in image inputs (see table \hyperref[tab:bspinflu]{\ref{tab:bspinflu}} and figure \hyperref[fig:bspextract]{\ref{fig:bspextract}} for an example). It has two approaches to generate data and mimic the Q-function; the first one is an \emph{experience training setting} which records and generates data during the training process for batch training. It records the state-action pairs and the resulting Q-values as `soft supervision labels' \cite[p. 1]{Liu2019utree} during training. In cases where the mimic learning model cannot be applied to the training process, the second approach can be used: \emph{active play setting}, which generates mimic data by applying the mature DRL to interact with the environment. Here, an online algorithm is required which uses stochastic gradient descent to dynamically update the linear models as more data is generated.

\begin{table}
    \centering
    \caption{Examples of feature influences in the Mountain Car and Cart Pole scenario, extracted by the LMUTs in \citeauthor{Liu2019utree} \cite{Liu2019utree}}
    \begin{tabular}{c|c|c}
         & Feature & Influence\\
         \hline
         Mountain & Velocity  & 376.86 \\
         Car & Position & 171.28 \\
         \hline
         & Pole Angle & 30541.54 \\
         Cart & Cart Velocity & 8087.68 \\
         Pole & Cart Position & 7171.71 \\
         & Pole Velocity At Tip & 2953.73 \\
    \end{tabular}
    \label{tab:bspinflu}
\end{table}

\begin{figure}[t]
    \centering
    \includegraphics[width=0.9\textwidth]{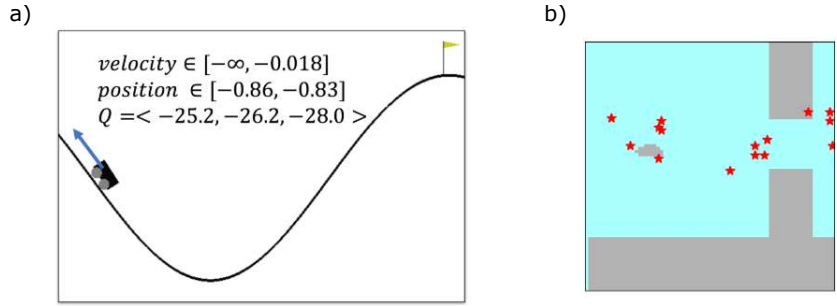}
    \caption{Examples of a) rule extraction, and b) super-pixels extracted by the LMUTs in \citeauthor{Liu2019utree} \cite{Liu2019utree}. a) Extracted rules for the mountain Cart scenario. Values at the top are the range of velocity and position and a Q vector ($Q_{move\_left,} Q_{no\_push}, Q_{move\_right}$) representing the average Q-value). In this example, the cart is moving to the left to the top of the hill. The car should be pushed left ($Q_{move\_left}$ is highest) to prepare for the final rush to the target on the right side. b) Super-pixels for the Flappy Bird scenario, marked by red stars. This is the first of four sequential pictures where the focus lies on the location of the bird and obstacles (i.e., pipes). In later pictures the focus would shift towards the bird's location and velocity.}
    \label{fig:bspextract}
\end{figure}

They evaluate the framework in three benchmark environments: Mountain Car, Cart Pole, and Flappy Bird, all simulated by the OpenAI Gym toolkit \cite{brockman2016openai}. Mountain Car and Cart Pole have a discrete action space and a continuous feature space, while Flappy Bird has two discrete actions and four consecutive images as inputs which result in 80x80 pixels each, so 6400 features. The LMUT method is compared to five other tree methods: a CART regression tree \cite{Loh2011}, M5 trees \cite{quinlan1992learning} with regression tree options (M5-RT) and with model tree options (M5-MT), and Fast Incremental Model Trees (FIMT, \cite{Ikonomovska2010}) in the basic version, and in the advanced version with adaptive filters (FIMT-AF). The two parameters \emph{fidelity} (how well the predictions of the mimic model match those from the mimicked model) and \emph{play performance} (how well the average return in the mimic model matches that of the mimicked model) are used as evaluation metrics. Compared to CART and FIMT (-AF), the LMUT model showed higher fidelity with fewer leaves. For the Cart Pole environment, LMUT showed the higest fidelity, while the M5 trees showed higher performance for the other two environments, although LMUT was comparable. Concerning the play performance, the LMUT model performs best out of all the models. This was likely due to the fact that, contrary to the LMUTs, the M5 and CART trees fit equally over the whole training experience which includes sub-optimal actions in the beginning of training, while the FIMT only adapts to the most recent input and thus cannot build linear models appropriately. In their work, this is represented by sorting the methods on an axis between `data coverage' (when the mimic model matches the mimicked model on a large section of the state space) and `data optimality' (when it matches the states most important for performance) with the LMUT at the, as they call it, `sweet spot between optimality and coverage' (p. 12, see also figure \hyperref[fig:sweetspot]{\ref{fig:sweetspot}}).

There is a similar, newer tree method that uses Soft Decision Trees (SDTs) to extract DRL polices \cite{coppens2019distilling}. This method was not presented in this paper because, for one thing, it is less versatile (not offering rule extraction, for example), and for another, it was not clear whether the SDTs actually adequately explained the underlying, mimicked policy for their used benchmark.

\begin{figure}[t]
    \centering
    \includegraphics[width=0.7\textwidth]{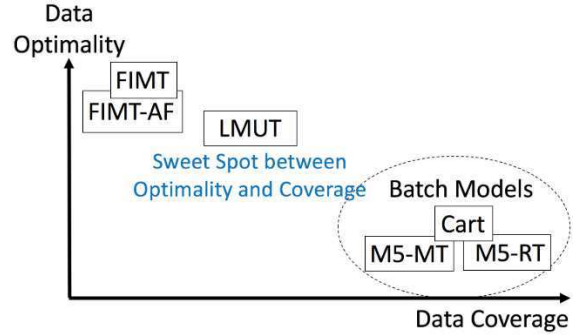}
    \caption{Placement of the different tree models on the axes data coverage vs. data optimality. Adapted from \citeauthor{Liu2019utree} \cite{Liu2019utree}.}
    \label{fig:sweetspot}
\end{figure}

\subsection{Method D: Explainable RL Through a Causal Lens}
\label{subsec:methd}
According to \citeauthor{MadumalCausal} \cite{MadumalCausal}, not only is it important for a RL agent to explain itself and its actions, but also to bear in mind the human user at the receiving end of this explanation. Thus, they took advantage of the prominent theory that humans develop and deploy causal models to explain the world around them, and have adapted a structural causal model (SCM) based on \citeauthor{Halpern2005} \cite{Halpern2005} to mimic this for model-free RL. SCMs represent the world with random exogenous (external) and endogenous (internal) variables, some of which might exert a causal influence over others. These influences can be described with a set of structural equations.

Since \citeauthor{MadumalCausal} \cite{MadumalCausal} focused on providing explanations for an agent's behaviour based on the knowledge of how its actions influence the environment, they extend the SCM to include the agent's actions, making it an \emph{action influence model}. More specifically, they offer `actuals' and `counterfactuals', that is, their explanations answer `Why?' as well as `Why not?' questions (e.g. `Why (not) action A?'). This is noticeable because, contrary to most XAI models, it not only considers actual events occured, but also hypothetical events that did not happen, but could have.

In more detail, the process of generating explanations consists of three phases; first, an action influence model in the form of a directed acyclic graph (DAG) is required (see figure \hyperref[SCM]{\ref{SCM}} for an example). Next, since it is difficult to uncover the true structural equations describing the relationships between the variables, this problem is circumvented by only approximating the equations so that they are exact enough to simulate the counterfactuals. In \citeauthor{MadumalCausal} \cite{MadumalCausal}, this is done by multivariate regression models during the training of the RL agent, but any regression learner can be used. The last phase is generating the explanations, more specifically, \emph{minimally complete contrastive explanations}. This means that, first, instead of including the vectors of variables of ALL nodes in the explanation, it only includes the absolute minimum variables necessary. Moreover, it explains the actual (e.g. `Why action A?') by simulating the counterfactual (e.g. `Why not action B?') through the structural equations and finding the differences between the two. The explanation can then be obtained through a simple NLP template (for an example of an explanation, again, see figure \hyperref[SCM]{\ref{SCM}}).

\citeauthor{MadumalCausal} \cite{MadumalCausal}'s evaluations of the action influence model show promising results; in a comparison between six RL benchmark domains measuring accuracy (`Can the model accurately predict what the agent will do next?') and performance (training time), the model shows reasonable task prediction accuracy and negligible training time. In a human study, comparing the action influence model with two different models that have learnt how to play Starcraft II ( a real-time strategy game), they assessed task prediction by humans, explanation satisfaction, and trust in the model. Results showed that the action influence model performs significantly better for task prediction and explanation satisfaction, but not for trust. The authors propose that, in order to increase trust, further interaction might be needed. In the future, advancements to the model can be made including extending the model to continuous domains or targeting the explanations to users with different levels of knowledge.

\begin{figure}[t]
\centering
\includegraphics[width=0.9\textwidth]{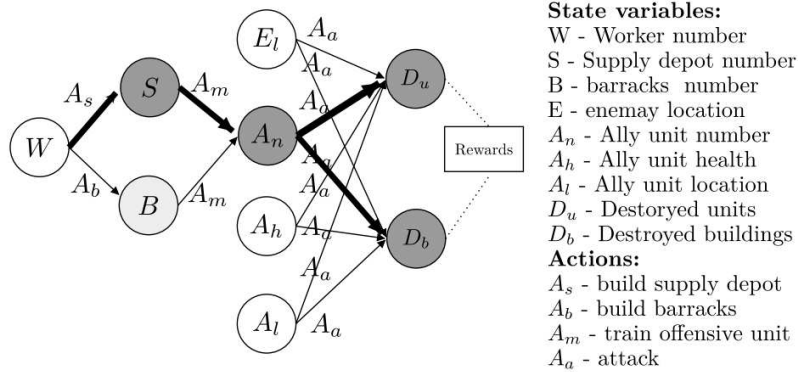}
\caption{Action influence graph of an agent playing Starcraft II, a real-time strategy game with a large state and action space, reduced to four actions and nine state variables for the purpose of generating the explanations. In this case, the causal chain for the actual action `Why $A_{s}$?' is shown in bold, and the chain for the counterfactual action `Why not $A_{b}$?' would be $B \to A_{n} \to [D_{u}, D_{b}]$. The explanation to the question `Why not build\_barracks ($A_{b})?$' would be `Because it is more desirable to do action build\_supply\_depot ($A_{s}$) to have more Supply Depots ($S$) as the goal is to have more Destroyed Units ($D_{u}$) and Destroyed buildings ($D_b$)'. Adopted from \citeauthor{MadumalCausal} \cite{MadumalCausal}.} \label{SCM} 
\end{figure}

\section{Discussion}

In this paper, inspired by the current interest in and demand for XAI, we focused on a particular field of AI: Reinforcement Learning. Since most XAI methods are tailored for supervised learning, we wanted to give an overview of methods employed only on RL algorithms, since, to the best of our knowledge, there is no work present at the current point in time addressing this.

First, we gave an overview over XAI, its importance and issues, and explained related terms. We stressed the importance of a uniform terminology and have thus suggested and defined a term to use from here on out. The focus, however, lay on collecting and providing an overview over the aforementioned XRL methods. Based on \textcite{Adadi2018}'s work, we have sorted selected methods according to the scope of the method and the time of information extraction. We then chose four methods, one for each possible combination of those categorizations, to be presented in detail.

Looking at the collected XRL methods, it becomes clear that post-hoc interpretability models are much more prevalent than intrinsic models. This makes sense, considering the fact that RL models were developed to solve tasks without human supervision that were too difficult for un-/supervised learning and are thus highly complex; it is, apparently, easier to simplify an already existing, complex model than it is to construct it to be simple in the first place. It seems that the performance-interpretability trade-off is present not only for the AI methods themselves, but also for the explainability models applied to them.

The allocation to global vs. local scope, however, seems to be more or less balanced. Of course, the decision to develop a global or a local method is greatly dependent on the complexity of the model and the task being solved, but one should also address the question if one of the two is more useful or preferable to human users. In \citeauthor{vanderwaacontr}'s study \cite{vanderwaacontr}, for example, `human users tend to favor explanations about policy rather than about single actions' (p. 1). 

In general, the form of the explanation and the consideration of the intended target audience is a very important aspect in the development of XAI/XRL methods that is too often neglected \cite{Abdul2018}. XAI methods need to exhibit \emph{context-awareness}: adapting to environmental and user changes like the level of experience, cultural or educational differences, domain knowledge, etc., in order to be more human-centric \cite{Adadi2018}.
The form and presentation of the explanation is essential as XAI `can benefit from existing models of how people define, generate, select, present, and evaluate explanations' \cite[p. 59]{Miller2019}. For example, research shows that (causal) explanations are contrastive, i.e., humans answer a `Why X?' question through the answer to the -often only implied- counterfactual `Why not Y instead?'. This is due to the fact that a complete explanation \emph{for} a certain event (instead of an explanation \emph{against} the counterevent) involves a higher cognititve load \cite{Miller2019}. Not only that, but a layperson also seems to be more receptive to a contrastive explanation, finding it `more intuitive and more valuable' \cite[p. 20]{Miller2019}).

Out of the papers covered in this work, we highlight \citeauthor{MadumalCausal}'s work \cite{MadumalCausal}, but also \citeauthor{sequeirainterestinglong} \cite{sequeirainterestinglong} and \citeauthor{vanderwaacontr} \cite{vanderwaacontr}; of all thirteen selected XRL methods, only five evaluate (non-expert) user satisfaction and/or utility of a method \cite{sequeirainterestinglong,juozapaitis2019explainable,vanderwaacontr,Fukuchi2017,MadumalCausal}, and only three of these offer contrastive explanations \cite{MadumalCausal,sequeirainterestinglong,vanderwaacontr}. So, of \emph{all} selected papers, only these free provide a combination of both, not only offering useful contrastive explanations, but also explicitly bearing in mind the human user at the end of an explanation.


\subsection{Conclusion}
For practical, legal, and psychological reasons, XRL (and XAI) is a quickly advancing field in research 
that has to address some key challenges to prove even more beneficial and useful.
In order to have a common understanding about the goals and capabilities of an XAI/XRL model, a ubiquitous terminology is important; due to this, we suggest the term \emph{interpretability} to be used from here on out and have defined it as `the ability to not only extract or generate explanations for the decisions of the model, but also to present this information in a way that is understandable by human (non-expert) users to, ultimately, enable them to predict a model's behaviour'. Different approaches are possible to achieve this interpretability, depending on the scope (global vs. local) and the time of information extraction (intrinsic vs. post-hoc). Due to the complexity of a RL model, post-hoc interpretability seems to be easier to achieve than intrinsic interpretability: simplifying the original model (for example with the use of a surrogate model) instead of developing a simple model in the first place seems to be easier to achieve, but comes at the cost of accuracy/performance.

What many models lack, however, is to consider the human user at the receiving end of an explanation and to adapt the model to them for maximum benefit. Research shows that contrastive explanations are more intuitive and valuable \cite{Miller2019}, and there is evidence that human users favor a global approach over a local one \cite{vanderwaacontr}. A context-aware system design is also important in order to cater to users with different characteristics, goals, and needs \cite{Adadi2018}. Especially considering the growing role of AI in critical infrastructures (for example analyzing and controlling power grids with models such as ARL \cite{Fischer2019arl,veith2019cpsanalysis}), where the AI model might have to act autonomously or in cooperation with a human user, being able to explain and justify the model's decisions is crucial.


To achieve this and be able to develop human-centered models for optimal and efficient human-computer interaction and cooperation, a bigger focus on interdisciplinary work is necessary, combining efforts from the fields of AI/ML, psychology, philosophy, and human-computer interaction.


\section{Acknowledgements}

This work was supported by the German Research Foundation under the grant GZ: JI 140/7-1. We thank our colleagues Stephan Balduin, Johannes Gerster, Lasse Hammer, Daniel Lange and Nils Wenninghoff for their helpful comments and contributions.


%
%
%
\bibliography{references.bib}
\end{document}